# Rational Counterfactuals

Tshilidzi Marwala

Joburg Univ.

South Africa

**Abstract.** This paper introduces the concept of rational countefactuals which is an idea of identifying a counterfactual from the factual (whether perceived or real) that maximizes the attainment of the desired consequent. In counterfactual thinking if we have a factual statement like: 'Saddam Hussein invaded Kuwait and consequently George Bush declared war on Iraq' then its counterfactuals is: 'If Saddam Hussein did not invade Kuwait then George Bush would not have declared war on Iraq'. The theory of rational counterfactuals is applied to identify the antecedent that gives the desired consequent necessary for rational decision making. The rational countefactual theory is applied to identify the values of variables *Allies*, *Contingency*, *Distance*, *Major Power*, *Capability*, *Democracy*, as well as *Economic Interdependency* that gives the desired consequent *Peace*.

## 1. Introduction

Rational decision making is important in many areas including economics, political science and engineering. Rational decision making involves choosing a course of action which maximizes the net utility. This paper explores counterfactual thinking in particular and introduces the theory of rational counterfactuals. The idea of using counterfactual thinking for decision making is an old concept that has been explored extensively by many researchers before (Lewis, 1973; Lewis, 1979).

In counterfactual thinking factual statements like: 'Saddam Hussein invaded Kuwait consequently George Bush declared war on Iraq', has a counterfactual: 'If Saddam Hussein did not invade Kuwait and George Bush would not have declared war on Iraq'.

Howlett and Paulus (2013) applied counterfactual thinking successfully in the problem of patient depression whereas Leach and Patall (2013) applied counterfactual reasoning for decision making on academic major. Celuch and Saxby (2013) applied successfully counterfactual reasoning in ethical decision making in marketing education whereas Simioni *et. al.* (2012) observed that multiple sclerosis decreases explicit counterfactual processing and risk taking in decision making. Fogel and Berry (2010) studied the usefulness of regret and counterfactuals on studying the disposition effect and individual investor decisions.

Johansson and Broström (2011) successfully applied counterfactual thinking in surrogate decision making. In this context incompetent patients have someone to make decisions on their behalf by having a surrogate decision maker make the decision that the patient would have made. Daftary-Kapur and Berry (2010) applied counterfactual reasoning on juror punitive damage award decision making while Shaffer (2009) studied decision theory, intelligent planning and counterfactuals to understand the limitation of the fact that Bayesian decision-theoretic framework does not sufficiently explain the causal links between acts, states, and outcomes in decision making.

In this paper, we develop a framework called a rational counterfactual machine which is a computational tool which takes in a factual and gives a counterfactual that is based on optimizing for the desired consequent by identifying an appropriate antecedent. This counterfactual is based on the learning machine and in this paper we choose the neuro-fuzzy network (Montazer *et. al.,* 2010; Talei *et. al.,* 2010) and an optimization technique and in this paper we choose simulated annealing (De Vicente *et. al.,* 2003; Dafflon *et. al.,* 2009). The rational counterfactual machine is applied to identify the antecedent that will give the consequent which is different from the consequent of the factual and the example that is applied in this paper is a problem of interstate conflict. This is done in a similar manner as it was done by Marwala and Lagazio (2004, 2011) as well as Tettey and Marwala (2004). The rational counterfactual machine is applied here to identify the values of antecedent variables *Allies*, *Contingency*, *Distance*, *Major Power*, *Capability*, *Democracy*, as well as *Economic Interdependency* that will give the consequent *Peace* given the factual statement.

This paper is organized as follows: the next section describes the notion of counterfactuals and then the following section describes the concept of rational counterfactual. Then a rational counterfactual theory is applied to the problem of interstate conflict.

## 2 Counterfactuals

Counterfactual thinking has been around for a very long time. Some of the thinkers who have dealt with the concept of counterfactuals include Hume (1748), Mill (1843), in Hegel's dialectic concept of thesis (i.e. factual), antithesis (i.e. counterfactual) and synthesis (Hegel, 1874) and Marx (1873). Counterfactual can be understood by breaking this word into two parts *counter* and *factual*. Factual is an event that has happened for example: Saddam Hussein invaded Kuwait and consequently George Bush declared war on Iraq. *Counter* means the opposite and in the light of the factual above: If Saddam Hussein did not invade Kuwait and consequently George Bush would not have declared war on Iraq. Of course counterfactual can be an imaginary concept and, therefore, the fundamental question that needs to be asked is: How do we know what would have happened if something did not happen? This paper addresses classes of problems where it is possible to estimate what might have happened and this procedure which we call a rational counterfactual machine is implemented via artificial intelligence techniques.

There are different types of counterfactuals and these include self/other as well as additive/subtractive. Additive and subtractive counterfactual is the case were the antecedent is either increased or decreased. One example of such will include: He drank alcohol moderately and consequently he did not fall sick. The counterfactual of this statement might be: He drank a lot alcohol and consequently he fell sick. The 'a lot' adds to the antecedent in the counterfactual.

There are a number of theories that have been proposed to understand counterfactuals and these include norm and functional theories (Birke *et. al.,* 2011; Roese, 1997). As described by Kahneman and Miller (1986) norm theory comprises a pairwise evaluation between a cognitive standard and an experiential outcome. Functional theory entails looking at how a counterfactual theory and its processes benefit people. Rational counterfactuals can be viewed as an example of the functional theory of counterfactuals.

Figure 1 indicates a factual and its transformation into a counterfactual. It indicates that in the universe of the counterfactuals that correspond to the factual there are many if not infinite number of counterfactuals. For example suppose we have a factual: Mandela opposed apartheid and consequently went to jail for 27 years. Its counterfactual can be: If Mandela did not oppose apartheid then he would not have gone to jail or If Mandela opposed apartheid gently he would not have gone to jail or If Mandela opposed apartheid peacefully he would not have gone to jail.

It is clear that there are multiple ways in which one can formulate counterfactuals for a given factual.

There a number of ways in which counterfactuals can be stated and this involves structural equations (Woodward, 2003; Woodward and Hitchcock, 2003). In the structural equation approach a counterfactual cab expressed as follows:

$$y = f(x_1, x_2, ..., x_n) \tag{1}$$

This expression can be read as: if it is the case that $x_1=X_1$, $x_2=X_2$,…, $x_n=X_n$ then it will be the case that $y=f(X_1, X_2,…, X_n)$. The usefulness of this approach will be apparent later in the paper when it is applied to modelling interstate conflict.

Figure 1 states that within the counterfactual universe there are group of counterfactuals that are called rational counterfactuals which are counterfactuals that are designed to maximize the attainment of particular consequences and these are called rational counterfactuals and are the subject of the next section.

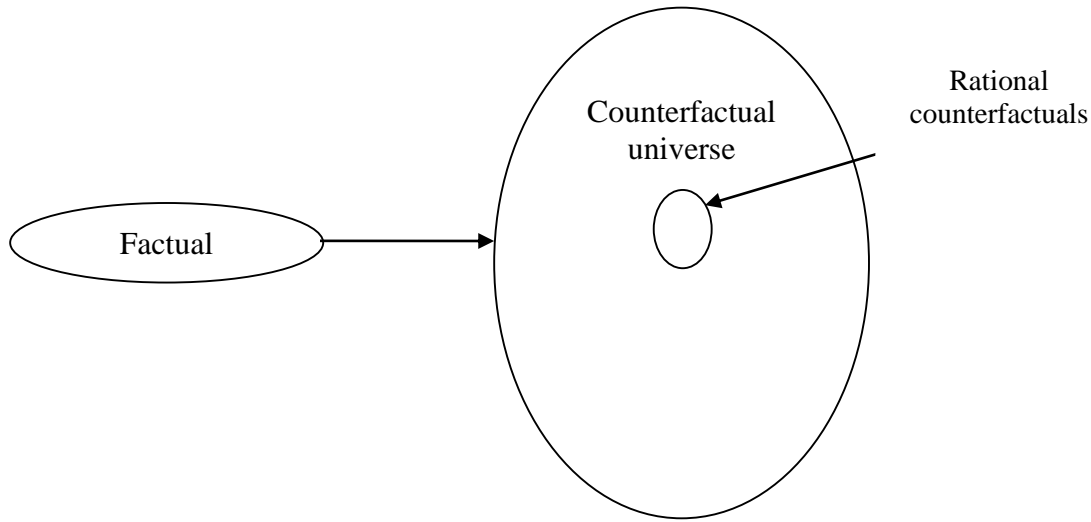

Figure 1 An illustration of a transformation of a factual into a counterfactual

## 3 Rational Counterfactuals

Now that we have discussed the concept of counterfactual, this section describes the concept of rational counterfactual and the corresponding machine for creating this concept. As shown in Figure 1, rational counterfactuals are those counterfactuals in the counterfactual universe corresponding to a given factual that maximizes a particular goal. There is a statement attributed to Karl Marx that states: "*The aim of a revolutionary is not merely to understand the world but to actually change it*". In this paper we therefore use counterfactual theory to solve practical problems and this is called the functional theory to counterfactual thinking. In this paper we also build what is known as a counterfactual machine, which is a computational system which gives a rational counterfactual whenever it is presented with a factual and a given problem domain. An illustration of a rational counterfactual machine is given in Figure 2. This figure shows that there are three objects in a rational counterfactual machine and these are: the factual which is antecedent into the rational counterfactual machine to give a rational counterfactual.

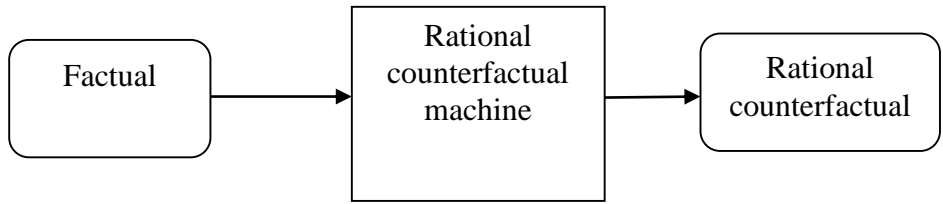

Figure 2 An illustration of a rational counterfactual machine

The rational counterfactual machine consists of a model that describes the structure and rules that define the problem at hand and a feedback which links the consequent (outcome) of the model and the antecedent. This model is shown in Figure 3.

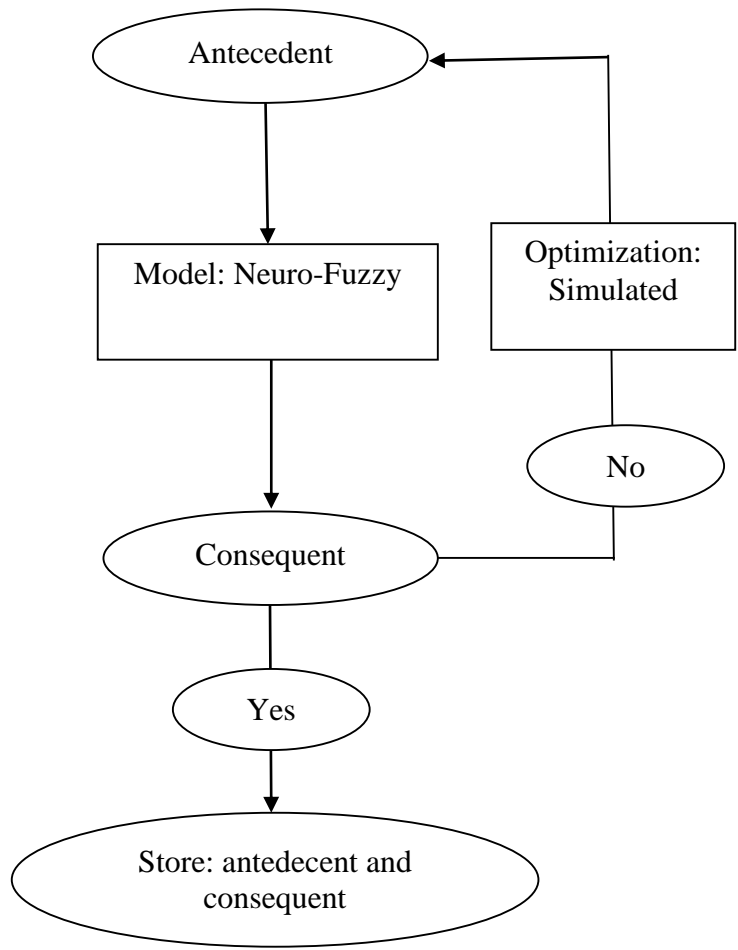

Figure 3 An illustration of a rational counterfactual machine

In this paper, we apply the problem of interstate conflict to illustrate the concept of rational counterfactuals. In this regard, we use neuro-fuzzy model to construct a factual relationship between the antecedent and the consequent. Then to identify the antecedent given the desired consequent and an optimization method. The objective function of the optimization problem is (Marwala and Lagazio, 2011):

$$error = \sum (y - t_d)^2 \qquad (2)$$

Here, $y$ is the neuro-fuzzy consequent and $t_d$ is the desired target consequent. Equation 2 is solved using the simulated annealing (Marwala and Lagazio, 2011). Equation 2 allows one to be able to construct a rational counterfactual which relates the identified antecedent (using equation 2) and the desired consequence.

## 4 Investigation and Results

The correlates of war (COW) data are used to produce training and testing sets. More information on this data set can be found in (Marwala and Lagazio, 2011). As in Marwala and Lagazio (2011) the training data set consists of 500 conflict- and 500 non-conflict cases, and the test data consists of 392 conflict data and 392 peace data. A balanced training set, with a randomly selected equal number of conflict- and non-conflict cases was chosen to yield robust classification and stronger comprehensions on the explanation of conflicts. The data were normalized to fall between 0 and 1. The antecedent variables were *Distance, Contiguity, Major Power, Allies*, *Democracy*, *Economic Interdependency*, and *Capability* and the consequent was either peace or war. In this regard and due to normalization two countries with the largest distance between their capitals will be assigned a value of Distance of 1 while the two countries with the shortest distance between their capitals will be assigned a Distance of 0. If both countries are superpower then they will be assigned variable Major Power of 1 while if only one is a value of 0.5 and if none are of 0. If two countries are not allies there are assigned a value of 0 while if they are of 1. If they are both democracies they will be assigned a value of 1 while if both are autocracy of 0. If the two countries share a border they will be assigned a Contiguity value of 1 while if they do not share a contiguity of 0. If the two countries have no economic interdependency the variable economic interdependency is 0 while if they have maximum economic interdependency recorded they are assigned a value of 1. For the maximum military capability the value is 1 while minimum is 0. Takagi-Sugeno neuro-fuzzy systems and simulated annealing were implemented to model militarized interstate dispute data. When these data are used in the modelling process a factual below is obtained:
 *If it is the case that Dis=0, C=1, MJ=0.4, A=0.1, D=0.3, EI=0.1, Cap=0.6) then it will be the case that Consequent=War.*
   Simulated annealing is used in a manner described in Figure 3 to identify the antecedent that would turn this factual into a counterfactual. In this regard the following rational counterfactual is identified that achieves peaceful outcome:
   *If it were the case that D=0.7, C=1, MJ=0.4, A=.8, D=0.3, EI=0.1, Cap=0.7) then it will be the case that Consequent =Peace.*
This counterfactual is deemed a rational counterfactual because it is formulated by identifying the antecedent which maximizes the attainment of a particular desired consequent and in this chapter peace.

## 5 Conclusions

This paper introduced rational countefactual which are counterfactuals that maximizes the attainment of the desired consequent. The theory of rational counterfactuals was applied to identify the antecedent that gives the desired consequent. The results obtained demonstrated the viability of a method f identifying rational counterfactuals.

## References

Birke D, Butter M, and Koppe T (eds) (2011) Counterfactual Thinking - Counterfactual Writing, Berlin, de Gruyter.

Celuch K and Saxby C (2013) Counterfactual Thinking and Ethical Decision Making: A New Approach to an Old Problem for Marketing Education. Journal of Marketing Education, 35 (2), pp. 155-167

Dafflon B, Irving J and Holliger K (2009) Simulated-annealing-based Conditional Simulation for the Local-scale Characterization of Heterogeneous Aquifers. J of Appl Geophys 68:60–70

Daftary-Kapur T and Berry M (2010) The effects of outcome severity, damage amounts and counterfactual thinking on juror punitive damage award decision making. American Journal of Forensic Psychology, 28 (1), pp. 21-4

De Vicente J, Lanchares J and Hermida R (2003) Placement by Thermodynamic Simulated Annealing. Phys Lett A 317:415–423

Fogel S O and Berry T (2010) The disposition effect and individual investor decisions: The roles of regret and counterfactual alternatives. Handbook of Behavioral Finance, pp. 65-80

Hegel G W F (1874) The Logic. Encyclopaedia of the Philosophical Sciences. 2nd Edition. London: Oxford University Press

Howlett J R and Paulus M P (2013) Decision-making dysfunctions of counterfactuals in depression: Who might I have been? Frontiers in Psychiatry, 4 (NOV), art. no. Article 143

Hume D (1748) An Enquiry concerning Human Understanding

Johansson M and Broström L (2011) Counterfactual reasoning in surrogate decision making - another look. Bioethics, 25 (5), pp. 244-249

Kahneman D, Miller D (1986) Norm theory: Comparing reality to its alternatives. Psychological Review 93 (2): 136–153

Leach J K and Patall E A (2013) Maximizing and Counterfactual Thinking in Academic Major Decision Making. Journal of Career Assessment, 21 (3), pp. 414-429

Lewis D (1973) Counterfactuals, Oxford: Blackwell

Lewis D (1979) Counterfactual Dependence and Time's Arrow, Noûs, 13: 455–76. Reprinted in his (1986a)

Marwala T and Lagazio M (2004) Modelling and Controlling Interstate Conflict. In: Proc of the IEEE Intl Joint Conf on Neural Nets:1233–1238

Marwala T and Lagazio M (2011) Militarized Conflict Modeling Using Computational Intelligence Springer-Verlag London

Marx K (1873) Capital Afterword to the Second German Edition, vol 1

Mill J S (1843) A System of Logic

Montazer G A, Saremi H Q and Khatibi V (2010) A Neuro-Fuzzy Inference Engine for Farsi Numeral Characters Recognition. Expert Syst with Appl 37:6327–6337

Roese N (1997) Counterfactual thinking. Psychological Bulletin 121 (1): 133–148

Shaffer M J (2009) Decision theory, intelligent planning and counterfactuals. Minds and Machines, 19 (1), pp. 61-92

Simioni S, Schluep M, Bault N, Coricelli G, Kleeberg J, du Pasquier R A, Gschwind M, Vuilleumier P, Annoni J-M (2012) Multiple Sclerosis Decreases Explicit Counterfactual Processing and Risk Taking in Decision Making. PLoS ONE, 7 (12), art. no. e50718

Talei A, Hock L, Chua C and Quek C (2010) A Novel Application of a Neuro-Fuzzy Computational Technique in Event-based Rainfall-Runoff Modeling. Expert Syst with Appl 37:7456–7468

Tettey T and Marwala T (2004) Controlling Interstate Conflict Using Neuro-Fuzzy Modeling and Genetic Algorithms. In: Proc of the 10th IEEE Intl Conf on Intelli Eng Syst:30–44

Tettey T and Marwala T (2007) Conflict Modelling and Knowledge Extraction Using Computational Intelligence Methods. In: Proc of the 11th IEEE Intl Conf on Intelli Eng Syst:161–166

Woodward J (2003) Making Things Happen: A Theory of Causal Explanation, Oxford: Oxford University Press

Woodward J and Hitchcock C (2003) Explanatory Generalizations. Part I: A Counterfactual Account, Noûs, 37: 1–24